%% file: ConstraintsVsLearning.tex
\documentclass[runningheads]{llncs}
\usepackage[autostyle]{csquotes}
\usepackage{amsmath}
\usepackage{amssymb}
\usepackage{algorithm2e}
\usepackage[capitalize]{cleveref}
\usepackage{graphicx}
\usepackage{subcaption}
\usepackage{booktabs}
\usepackage{tikz}
\usepackage{pgfplots}
\usepackage{cite}
\usepackage{tikz}
\usetikzlibrary{positioning}

\begin{document}
\title{Comparing Heuristics, Constraint Optimization, and Reinforcement Learning for an Industrial 2D Packing Problem}
\titlerunning{Solving an Industrial 2D Packing Problem}
%
\author{Stefan Böhm\inst{1,2}\and
Martin Neumayer\inst{3}\and
Oliver Kramer\inst{1}\and
Alexander Schiendorfer\inst{4}\and
Alois Knoll\inst{2}}
\authorrunning{S. Böhm et al.}
%
\institute{Research \&\ Development Department, Technische Hochschule Rosenheim, Rosenheim, Germany\\
\email{(stefan.boehm, oliver.kramer)@th-rosenheim.de} \and
Chair of Robotics, Artificial Intelligence and Real-time Systems, Technische Universität München, München, Germany\\
\email{knoll@mytum.de}\and
Institut für nachhaltige Energieversorgung, Rosenheim, Germany\\
\email{martin.neumayer@inev.de}\and
Research Institute AImotion Bavaria, Technische Hochschule Ingolstadt, Ingolstadt, Germany\\
\email{Alexander.Schiendorfer@thi.de}}
\maketitle              
\begin{abstract}
Cutting and Packing problems are occurring in different industries with a direct impact on the revenue of businesses. Generally, the goal in Cutting and Packing is to assign a set of smaller objects to a set of larger objects. To solve Cutting and Packing problems, practitioners can resort to heuristic and exact methodologies. Lately, machine learning is increasingly used for solving such problems. This paper considers a 2D packing problem from the furniture industry, where a set of wooden workpieces must be assigned to different modules of a trolley in the most space-saving way. We present an experimental setup to compare heuristics, constraint optimization, and deep reinforcement learning for the given problem. The used methodologies and their results get collated in terms of their solution quality and runtime. In the given use case a greedy heuristic produces optimal results and outperforms the other approaches in terms of runtime. Constraint optimization also produces optimal results but requires more time to perform. The deep reinforcement learning approach did not always produce optimal or even feasible solutions. While we assume this could be remedied with more training, considering the good results with the heuristic, deep reinforcement learning seems to be a bad fit for the given use case.

\keywords{Cutting and Packing Problem  \and Constraint Optimization \and Machine Learning.}
\end{abstract}
\input{sections/Motivation}
\input{sections/ProblemDefinition}
\input{sections/RelatedWork}

\input{sections/Implementation/Implementation}
\input{sections/Evaluation/Evaluation}
\input{sections/Conclusion}
%
%
\bibliographystyle{splncs04}
\bibliography{bib}

\end{document}

%% file: sections/Motivation.tex
\section{Motivation}
Cutting and Packing (C\&P) problems are ubiquitous in many areas of application like processing wood, metal, glass, or fabric. Despite the different fields of application, C\&P problems share a common structure: There is a set of small objects and a set of large objects. Some or all of the small items are then assigned to one of the large objects, such as the small objects lie within the large objects and the small objects do not overlap \cite{wascher2007improved}. Saving resources like time by the efficient sorting of workpieces, solving C\&P problems have a direct impact on the revenue of many businesses. However, many C\&P problems might take too long to be solved exactly. Consequently, C\&P problems are extensively studied in literature and for many of these problems a variety of algorithms exist, see \cite{rao2019engineering} and the references therein.

To solve C\&P problems practitioners can choose between heuristic algorithms and exact methods among others. While heuristics offer fast but possibly suboptimal results, exact methods promise an optimal solution with the disadvantage of requiring more time. Lately, machine learning algorithms are proposed as an alternative for solving complex control and optimization tasks \cite{bengio2018machine}. Especially when the runtime of exact algorithms increases, e.g. due to the size of the search space, machine learning algorithms seem promising as they can \enquote{replace some heavy computations by a fast approximation}\cite{bengio2018machine}. Yet, studies comparing runtime and solution quality are often missing. Therefore, practitioners are left alone with choosing a suitable algorithm for a specific use case. Additionally, the no free lunch theorem \cite{wolpert1997no} indicates that there is no single algorithm that can solve all optimization problems efficiently. Instead, Wolpert and Macread state that any two algorithms will perform equally well when their performance is averaged across all possible problems \cite{wolpert2005coevolutionary}. Thus, the search for efficient algorithms for special problem classes is relevant, not only to practitioners but also to researchers.

To this point in this paper, we compare heuristics, constraint optimization, and deep reinforcement learning for a packing problem found in the furniture industry. In the presented problem wooden workpieces have to be packed into a trolley used for transport in the most space-saving way. By comparing the different methods in terms of runtime and solution quality we provide insight and assist practitioners in selecting an approach.

The remainder of this paper is structured as follows: \cref{problem} discusses the use case and our assumptions alongside a formal description of the problem considered. \cref{related-work} provides related work in the field of heuristics, constraint programming, and deep reinforcement learning for packing problems. In \cref{implementation} we present implementation that is evaluated against random and real data in \cref{evaluation}. \cref{conclusion} summarizes our findings and concludes this paper with recommendations for practitioners.

%% file: sections/ProblemDefinition.tex
\section{Problem Definition}
\label{problem}

In this section, we present an industrial packing problem as the use case for comparing different approaches. First, we introduce the trolley at the center of our use case. We point out our assumptions when considering the problem of packing workpieces into the trolley and provide a formal problem description.

\subsection{Use Case: Trolley}
\begin{figure*}
	\centering
	\begin{subfigure}[t]{0.5\textwidth}
		\centering
		\includegraphics[height=1.8in]{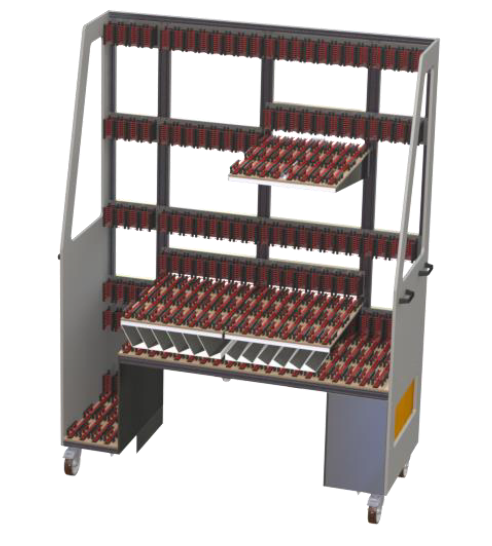}
		\caption{3D model}
		\label{fig:container_real}
	\end{subfigure}%
~
	\begin{subfigure}[t]{0.5\textwidth}
		\centering
		\includegraphics[height=1.8in]{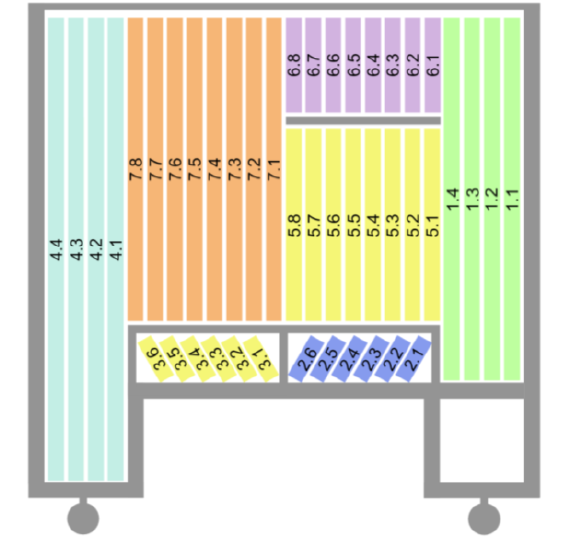}
		\caption{Schematic representation of modules and slots}
		\label{fig:container_schema}
	\end{subfigure}
	\caption{Representations of the trolley}
	\label{fig:container_figures}
\end{figure*}

Trolleys, like depicted in \cref{fig:container_real}, are considered standard means of transportation for workpieces in the furniture industry. The implementation presented in \cref{fig:container_real} is equipped with a car battery and a single-board Raspberry Pi computer. This computer enables the trolley to communicate wirelessly with machines, control and information systems on the one hand, on the other hand, with business process, thus realizes the idea of industrial internet of things device \cite{Sisinni2018industrial}. A human operator can communicate with the trolley via a pick-by-light system and a tablet application. While the trolley can be moved by a human operator pulling the handles, the cutout at the bottom also enables transport via autonomous guided vehicle (AGV).

\cref{fig:container_schema} schematically demonstrates the structure of the trolley: It is made of \textit{slots}, displayed as colored rectangles. A slot is a three-dimensional space in the trolley able to fit one \textit{workpiece}  at a time. Slots with the same dimensions are subsumed as a \textit{module}. The number of slots in a module is also called its \textit{capacity} in the following. As an example, consider the rightmost slot 1.1 in \cref{fig:container_schema}: It is part of module 1 which has a capacity of 4.  

\begin{figure*}
	\centering
	\includegraphics[height=2.8in]{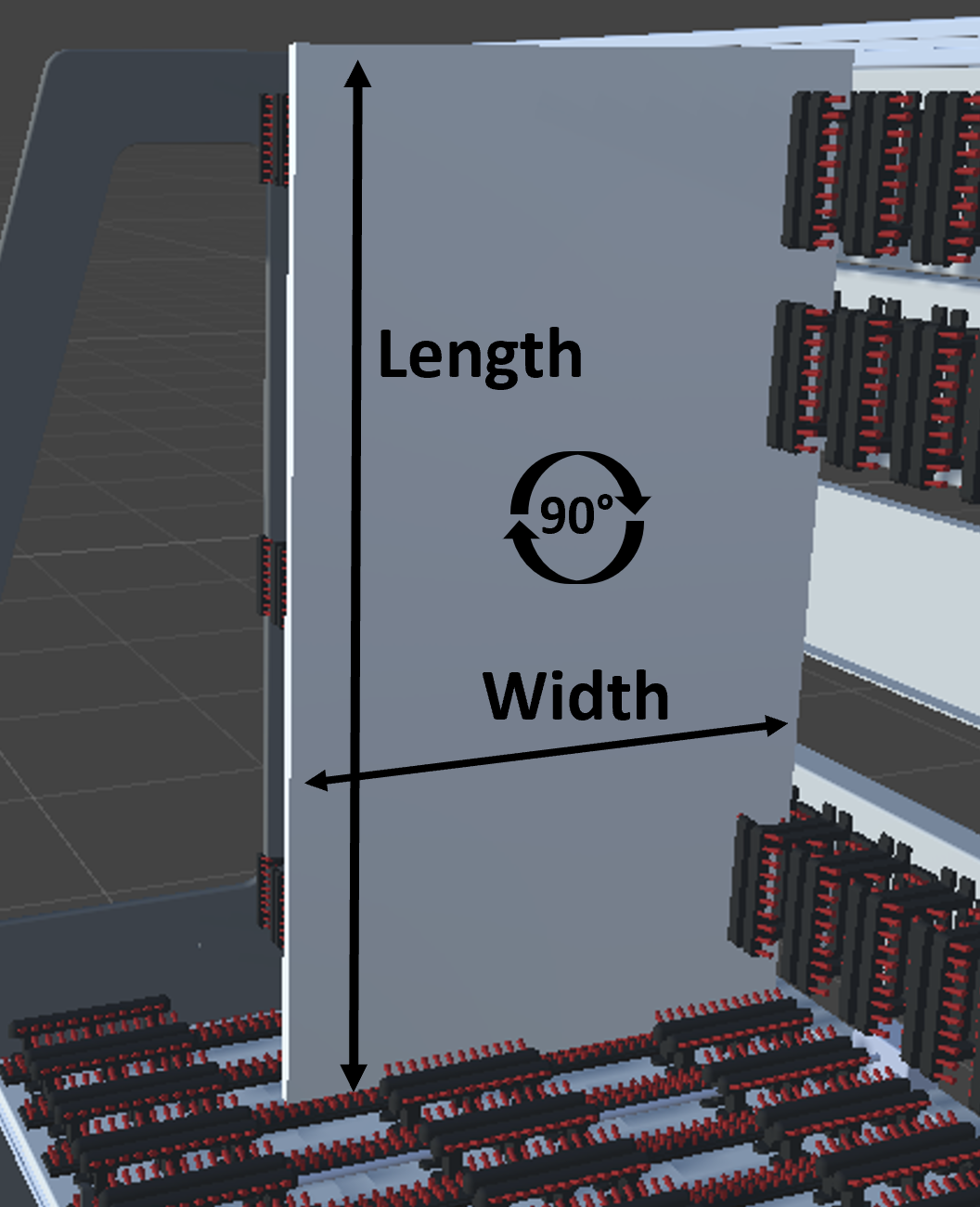}
	\caption{Possible packing and rotation option of a part. We denote the longer side of the part as length and the shorter side as width. The thickness is ignored.}
	\label{fig:Board_geometry}
\end{figure*} 

\subsection{Assumptions and Limitations}
\label{problem_assumptions}
Now we consider the problem of packing workpieces into the aforementioned trolley as shown in \cref{fig:Board_geometry} with the following assumptions and limitations:

\begin{enumerate}
	\item We assume that the specified modules can accommodate all workpieces. We do not consider the case where the number of workpieces exceeds the modules' capacity. Further, we do not consider cases where we have to decide which workpieces to pack. 
	\item A slot can only hold one workpiece at a time to ensure clear identification of the workpiece when using the pick-by-light system.
	\item The thickness of workpieces and slots is ignored as the trolley is equipped with brushes that allow holding parts with typical thickness used in the furniture industry.
	\item Rotation of workpieces by 90° as shown in \cref{fig:Board_geometry} is allowed.
	\item We assume that assigning a slot is trivial once a fitting module with sufficient capacity is found. Therefore, we only consider the assignment of modules, not concrete slots. 
\end{enumerate}

\subsection{Formal Description}
\label{problem_formal}
Let us introduce the problems in a formal way. We consider

\begin{itemize}
	\item a set of rectangular parts $p \in P$, possessing a longer side called length, $l_p$, and a shorter side called width, $w_p$,
	\item a set of modules $m \in M$, possessing length $l_m$, width $w_m$ and capacity $c_m$,
	\item a set of assignment variables $x_{pm}$, with $m \in M, p \in P$, where $x_{pm} = 1$, if part $p$ is assigned to module $m$ and $x_{pm} = 0$, otherwise.
\end{itemize}

Thus, a possible problem formulation is:

\begin{alignat}{2}
	\textrm{minimize }& \sum_{p \in P, m \in M} o_{pm} \cdot x_{pm} \label{eq:obj}\\
	\textrm{subject to }& o_{pm} = l_m \cdot w_m - l_p \cdot w_p \label{eq:costs}\\
	&x_{pm} = 1 \Rightarrow \nonumber \\
	& \qquad \qquad (l_p \le l_m \wedge w_p \le w_m) \vee \nonumber \\ 
	& \qquad \qquad (l_p \le w_m \wedge w_p \le l_m) && \qquad \forall p \in P, \forall m \in M \label{eq:length_width}\\
	&\sum_{p \in P} x_{pm} \le c_m && \qquad \forall m \in M \label{eq:capacity} \\
	&\sum_{m \in M} x_{pm} = 1 && \qquad \forall p \in P \label{eq:one_asignment}\\
	& x_{pm} \in \{0,1\} && \qquad \forall p \in P, \forall m \in M
	\label{eq:bounds}
\end{alignat}

Our objective is to minimize the wasted space. Therefore, we calculate the wasted space $o_{pm}$ for assigning workpiece $p$ to module $m$ in \cref{eq:costs}: The area of the workpiece is subtracted from the area of the module. Then we minimize the sum of wasted space overall assignments in \cref{eq:obj}. \cref{eq:length_width} states a workpiece's length and width must not exceed the length and width of the module it is placed in. \cref{eq:length_width} also considers rotation as the workpiece's length and width might be exchanged. \cref{eq:capacity} ensures that a module's capacity is not exceeded. Lastly, \cref{eq:one_asignment} and \cref{eq:bounds} guarantee the assignment of every workpiece to exactly one module.

%% file: sections/RelatedWork.tex
\section{Background and Related Work}
\label{related-work}

\subsection{Heuristics for Bin Packing Problems}
The problem at hand resembles the Bin Packing Problem. Therefore, we can adapt and reuse well-known heuristics, e.g. from the classical one-dimensional Bin Packing Problem with unlimited bins. In contrast to the problem considered here, a new bin is opened whenever no bin can accommodate a given item. Although a large number of heuristics exist \cite{Coffman1996approximation}, we will concentrate on one of the most common and relevant ones: Best Fit (BF). The Best Fit heuristic places an item in a bin in which the item still fits \cite{Coffman1996approximation}. Theoretical analysis show that the Best Fit heuristic has a worst-case performance of $\lfloor 1.7 \cdot \text{OPT} \rfloor$ bins, where $\text{OPT}$ is the optimal number of bins \cite{dosa2013first,dosa2014optimal}.

\subsection{Constraint Programming}
Constraint Programming is a declarative approach to programming: A user describes a problem by specifying  requirements or constraints \cite{apt2003principles}. Formally, Constraint Optimization Problem (COP) are often descibed as a tuple $\left\langle X, D, C, f \right\rangle$, where
\begin{itemize}
	\item $X = \lbrace x_1, \dots, x_n \rbrace $ is a finite set of variables.
	\item $D = \lbrace D_1, \dots, D_n \rbrace$ denotes the finite sets of domains for the variables in $X$. $D_i$ corresponds to the set of possible values for $x_i$.
	\item $C$ is a finite set of constraints specifying allowable combinations of values over subsets of $X$.
	\item $f: \left[X \rightarrow D \right] \rightarrow \mathbb{R} $, that maps an assignment of values to a real number representing the cost of this assignment \cite{russell2012artificial}.
\end{itemize}

A solution is provided by a constraint solver. A constraint solver is a piece of software that often relies on backtracking search, inference, and more specialized algorithms depending on the problem description. Intuitively, a solution is an assignment of values for all variables in $X$ that is consistent with the respective domains and satisfies all constraints in $C$. An optimal solution further minimizes the function $f$ \cite{russell2012artificial}.

Constraints involving an arbitrary number of variables (but not necessarily all variables) are termed global constraints. Instead of using the general-purpose methods mentioned before, solvers devise optimized algorithms to handle frequently occurring problems \cite{russell2012artificial}.

\subsection{Deep Reinforcement Learning} 
\label{Deep Reinforcement Learning}
Sutton and Barto \cite{Sutton.2018} describe reinforcement learning as methods in which an agent learns to take actions based on his observation of the environment and the received reward signal.

That is why Sutton and Barto consider reinforcement learning as its own machine learning paradigm and describe it with three main elements. The first element is the policy which defines the learning agent's behavior. The policy is depending on a reward signal, which is the second element. This reward signal defines the goals of the training. Therefore, it describes how successful actions are. The third element is a value function. 
It estimates the value of a given state, also considering future states. Using the value function, the agent can maximize its cumulative reward and overcome shortsighted decisions.

Mnih et al. present in \cite{Mnih.2015} a novel reinforcement learning approach called deep Q-network (DQN). This approach uses a combination of reinforcement learning and an artificial neuronal network. Hereby, the neuronal network approximates the optimal action-value function $Q\textsuperscript{*}(s,a)$, which is the agent's maximum sum of rewards. However, because DQN tends to overestimate the action value, van Hasselt et al. \cite{vanHasselt.22.09.2015} presents an improved version of DQN called Double DQN (DDQN) which uses the Double Q-learning approach to solve the overestimation problem. To avoid that a neural network has to learn the effect of each action for each observation, \cite{pmlr-v48-wangf16} presents the dueling architecture for modern reinforcement learning approaches. This architecture has two streams, one for the value- and the other for the advantage function. Both streams are combined in an aggregating layer to estimate the state-action value function $Q$. This has proven to be useful in states where the agent's action does not affect future observation.

DQN is already used for solving bin packing problems: In \cite{Verma.01.07.2020} DQN is used to produce decisions for a robot arm to solve 3D bin packing problems. It is shown that the DQN approach outperforms specialized greedy heuristics in this setting. Lu Duan et al. \cite{Duan.17.04.2018} presents a study about other Reinforcement Learning approaches besides DQN for a 3D bin packing problem. The paper aims to find the best possible approach to pack a fixed number of cuboid-shaped items into a rectangular bin with minimum wasted space.

%% file: sections/Implementation/Implementation.tex
\section{Implementation}
\label{implementation}
In this section, we present the implementation of the best fit heuristic, constraint optimization, and the deep reinforcement learning algorithm. To enable a comparison all three algorithms were implemented in Python. For a better understanding of the implementation and results, we have published the program at \cite{GitHub.09.02.2021}. The program provides two possibilities for evaluation. One is to evaluate randomly generated data, the other contains a predefined data set from real parts. The number of parts to be evaluated can be defined in both cases manually. Assumption 1 from \cref{problem_assumptions} states that the number of necessary containers for the evaluation is already known. Therefore, the program predefines how many containers are necessary for a best fit packing. The program evaluates the data with all three algorithms and the results for runtime and solution quality, which is the wasted space, are plotted. For each packed part, the capacity in the corresponding module is reduced by one. 

\subsection{Best Fit Heuristic}
The best fit heuristic is a greedy algorithm based on \cref{eq:obj} to \cref{eq:capacity}. 
Pseudocode for the best fit heuristic is presented in \cref{alg:best_fit}.
Given a part to pack, the heuristic first finds all fitting modules with a capacity greater than 0.
Then the wasted space for each of these fitting modules is calculated according to \cref{eq:costs}.
Lastly, the module with the least wasted space is returned as the best fit.
Note that a faster variant of the algorithm can be implemented using binary trees \cite{johnson1974fast}.

\begin{algorithm}
	\DontPrintSemicolon
	\KwData{\\
		$part$ : the part to be packed\\
		$modules$: a list of available modules}
	\BlankLine
	\For{$module \in modules$}
	{
		\If{part fits in module and module.capacity $ > 0$}{
			Calculate the wasted space for packing part into module according to \cref{eq:costs}\;
		}
	}
	return the module that minimizes the wasted space\;
	\caption{Pseudocode for the best fit heuristic}
	\label{alg:best_fit}
\end{algorithm}

\subsection{Constraint Programming in MiniZinc}
We choose MiniZinc \cite{Nethercote2007MiniZinc} as a modeling language for the packing problem as a COP, as it is open-source and supports a range of solvers. This allows us to test the performance of different solvers and also makes it a reasonable choice for practitioners. Furthermore, MiniZinc offers global constraints for bin packing problems, promising time savings in modeling, and reasonable performance in solving the problem described in \cref{problem}. After describing the problem formally in \cref{problem_formal}, modeling as a COP in MiniZinc is straightforward. Therefore, we tested a first constraint model that closely resembled the formal problem description. This first model calculated feasible allocations as a 2D matrix of boolean assignment variables. Finally, we settled for a second model that uses the \texttt{bin\_packing\_capa} global constraint to find an array of feasible allocations, as this second model outperforms the first in terms of runtime. As a constraint solver, we choose Gurobi \cite{gurobi2020}. While Gecode \cite{gecode2020} can solve small instances equally fast, Gurobi performs better on instances larger than 15 parts. We further use search heuristics to search for the allocation with the smallest domain size first and to assign the smallest domain value first. Adding additional constraints for symmetry breaking \cite{fahle2001symmetry} worsened the performance, thus it is omitted.

\subsection{Deep Reinforcement Learning}
Because of the mentioned advantages in \cref{Deep Reinforcement Learning} for bin packing problems and its versatile and generic capabilities without the need for labeling training data manually \cite{Mnih.19.12.2013, Mnih.2015} we have chosen the DQN algorithm. Specifically we choose the DDQN  \cite{vanHasselt.22.09.2015} setup with dueling network architecture \cite{pmlr-v48-wangf16} from the Keras-RL2 python package. The DDQN has an input layer $I$ with 20 neurons, 3 hidden layers $H$ of 32 neurons each, and 6 neurons as output layer $Q$. The described DDQN structure and its hyperparameters are shown in \cref{fig:neuronal_network} and \cref{table:hyperparameters}. 
The observed length and width of one board and each of the 6 modules, as well as all free capacities, are fed numerically into the input layer. The reward system is based on the Best Fit Heuristic: If the DDQN packs the observed board in the best fitting module it receives a reward of 1. A negative reward of -1 is given for each incorrectly selected module. 
\input{sections/Implementation/NeuronalNet_Hyperparameters}

%% file: sections/Implementation/NeuronalNet_Hyperparameters.tex
\begin{figure*}
	\centering

\tikzset{%
   input neuron/.style={
	fill=green!50,    
    circle,
    draw,
    minimum size=0.75cm
   },
   hidden neuron/.style={
   	fill=blue!50,
    circle,
    draw,
    minimum size=0.75cm
   },
   output neuron/.style={
	fill=orange!50,    
    circle,
    draw,
    minimum size=0.75cm
   },
   neuron missing/.style={
	fill=none,   
    draw=none,
    scale=4,
    text height=0.333cm,
    execute at begin node=\color{black}$\vdots$
   },
}

\resizebox{\textwidth}{!}{%
\begin{tikzpicture}[x=1.25cm, y=1.5cm, >=stealth]

\foreach \m/\l [count=\y] in {1,2,missing,3}
  \node [input neuron/.try, neuron \m/.try] (input-\m) at (0,1.75-\y) {};

\foreach \m [count=\y] in {1,2,missing,3,4}
  \node [hidden neuron/.try, neuron \m/.try ] (hidden1-\m) at (2.25,2.3-\y) {};
  
\foreach \m [count=\y] in {1,2,missing,3,4}
  \node [hidden neuron/.try, neuron \m/.try ] (hidden2-\m) at (4.5,2.3-\y) {};
  
\foreach \m [count=\y] in {1,2,missing,3,4}
  \node [hidden neuron/.try, neuron \m/.try ] (hidden3-\m) at (6.75,2.3-\y) {};

\foreach \m [count=\y] in {1,missing,2}
  \node [output neuron/.try, neuron \m/.try ] (output-\m) at (9,1.3-\y) {};

\foreach \l [count=\i] in {1,2,20}
  \draw [<-] (input-\i) -- ++(-1,0)
    node [above, midway] {$I_{\l}$};

\foreach \l [count=\i] in {1,2,31,32}
  \node [above] at (hidden1-\i.north) {$H_{\l}$};
  
\foreach \l [count=\i] in {1,2,31,32}
  \node [above] at (hidden2-\i.north) {$H_{\l}$};
  
\foreach \l [count=\i] in {1,2,31,32}
  \node [above] at (hidden3-\i.north) {$H_{\l}$};
  
\foreach \l [count=\i] in {1,6}
  \draw [->] (output-\i) -- ++(1,0)
    node [above, midway] {$Q_{\l}$};

\foreach \i in {1,...,3}
  \foreach \j in {1,...,4}
    \draw [->] (input-\i) -- (hidden1-\j);

\foreach \i in {1,...,4}
  \foreach \j in {1,...,4}
    \draw [->] (hidden1-\i) -- (hidden2-\j);
    
\foreach \i in {1,...,4}
  \foreach \j in {1,...,4}
    \draw [->] (hidden2-\i) -- (hidden3-\j);
    
\foreach \i in {1,...,4}
  \foreach \j in {1,...,2}
    \draw [->] (hidden3-\i) -- (output-\j);

\foreach \l [count=\x from 0] in {Input, First Hidden, Second Hidden, Third Hidden, Ouput}
  \node [align=center, above] at (\x*2.25,2) {\l \\ layer};
  
\end{tikzpicture}
}	
	
	\caption{Visualisation of the DDQN structure}
	\label{fig:neuronal_network}
\end{figure*}
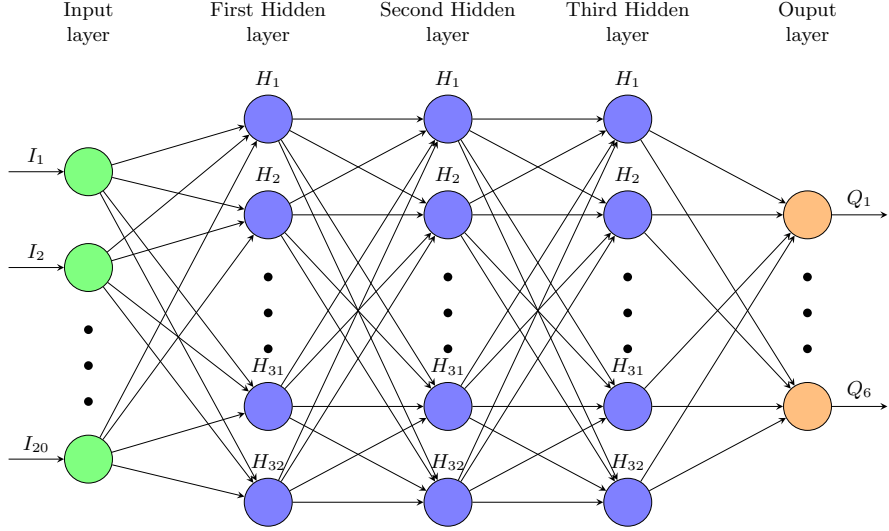

\begin{table}[]
\centering
\begin{tabular}{lr}
\hline
\multicolumn{1}{|l|}{Hyperparameter}                                                            & \multicolumn{1}{l|}{Value}                                            \\ \hline
\multicolumn{1}{|l|}{memory\_Limit} 
& \multicolumn{1}{r|}{1000} 											\\ \hline				
\multicolumn{1}{|l|}{np\_steps\_warmup}                                                         & \multicolumn{1}{r|}{500}                                              \\ \hline
\multicolumn{1}{|l|}{target\_model\_update}                                                     & \multicolumn{1}{r|}{1e-2}                                             \\ \hline
\multicolumn{1}{|l|}{enable\_dueling\_network}                                                  & \multicolumn{1}{r|}{True}                                             \\ \hline
\multicolumn{1}{|l|}{dueling\_type}                                                             & \multicolumn{1}{r|}{avg}                                              \\ \hline
\multicolumn{1}{|l|}{enable\_double\_dqn}                                                       & \multicolumn{1}{r|}{True}                                             \\ \hline
\multicolumn{1}{|l|}{policy}                                                                    & \multicolumn{1}{l|}{BoltzmannQPolicy}                                 \\ \hline                                                 
\end{tabular}
\caption{Hyperparameters used for the training of the DDQN}
\label{table:hyperparameters}
\end{table}

%% file: sections/Evaluation/Evaluation.tex
\section{Evaluation}
\label{evaluation}
In this section, we compare the three presented approaches experimentally. Therefore, we run several simulations with an increasing number of either randomly generated or real furniture parts. To compare the approaches we measure the following properties:  the runtime and the sum of the wasted space according to \cref{eq:obj}.
Measuring these properties with an increasing number of parts allows us to conclude the scalability of the approaches. This procedure also allows comparing approaches that can solve larger instances to approaches that can not.

The following hypotheses are tested with our evaluations:
\begin{itemize}
	\item Hypothesis 1: As the Constraint Optimization approach is our baseline for optimal solution quality, it will always provide a feasible and optimal solution. However, this approach will require more time compared to Deep Reinforcement Learning and the Best Fit heuristic.
	\item Hypothesis 2: If the Best Fit heuristic or Deep Reinforcement Learner generates a feasible solution, it will take less runtime than the Constraint Optimization approach.
	\item Hypothesis 3: With an increasing number of parts the Deep Reinforcement Learning and the Best Fit heuristic may maneuver into a dead end because they only consider one part at a time and are not able to backtrack. Therefore, not every solution provided may be optimal or even feasible.
\end{itemize}

\subsection{Experimental Setup}
To verify our hypotheses, we run a set of experiments where we steadily increase the number of parts. This methodology allows us to draw conclusions about the scalability of the algorithms. Furthermore, this methodology points out clearly if and at what point algorithms are not able to create feasible solutions anymore. Once we exceed the trolley's capacities, i.e. no feasible solution exists, we increase the capacities as if there were several trolleys.

We start our experiments by drawing one part and add it to the parts to pack. Then the minimum number of trolleys is determined, starting with one. Next, we run our algorithms using the parts to pack and the number of trolleys as inputs. After the execution, we check the found solution for feasibility and save the measured properties, i.e. runtime and wasted space. Unless we have reached the maximum number of parts, we draw a new part and add it to the parts to pack. We repeat determining the number of trolleys to run our algorithms.

We exclude infeasible solutions from the results. A solution can be infeasible for two reasons: It might either allocate a module that is too small to fit the given part (violating \cref{eq:length_width}) or exceed the module's capacity (violating \cref{eq:capacity}). Assuming the used algorithms are deterministic, an algorithm that can not provide a solution for a given set of parts, may not be able to solve the following set of parts either, if it includes the prior set of parts. However, once the minimum number of trolleys is increased, the algorithm can produce a feasible solution again.

For our experiments, we use both, parts of a real furniture product, i.e., a fitted kitchen and randomly generated parts. Since real product parts often resemble each other and follow certain standards, the predefined real product data serve as a reference for the solution quality of the algorithms under real conditions. On the other hand, the randomly generated data shows the solution quality for the whole spectrum of possible parts and thus the general solution quality.

\subsection{Experimental Results}

\begin{figure*}
\input{sections/Evaluation/ExperimentalSetup}
\label{fig:RandomPartGraphs}
\caption{Graphs a to d show the results of the three evaluated algorithms Constraint Programming (blue), Best Fit Heuristics (green), and DDQN (black) for randomly generated data.}
\end{figure*}
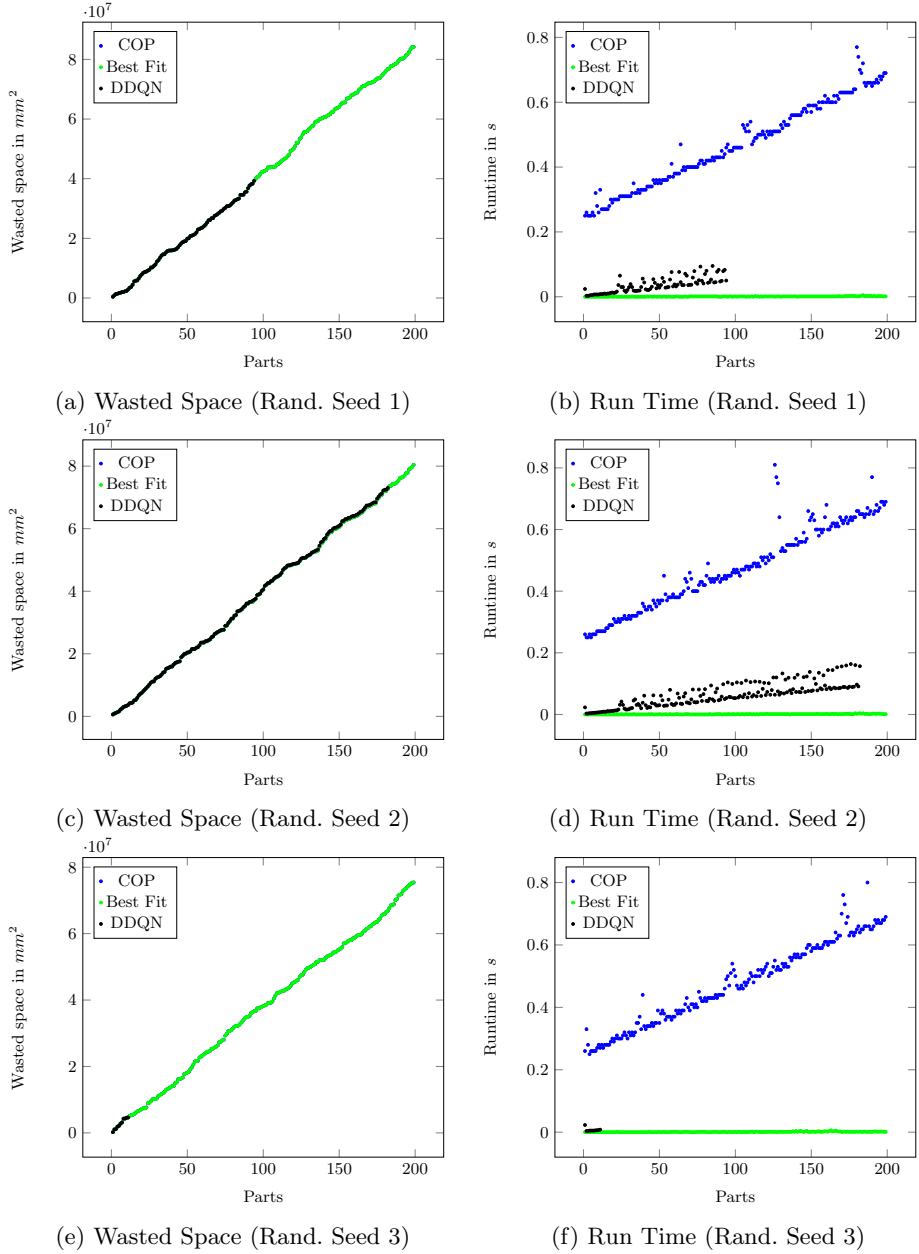

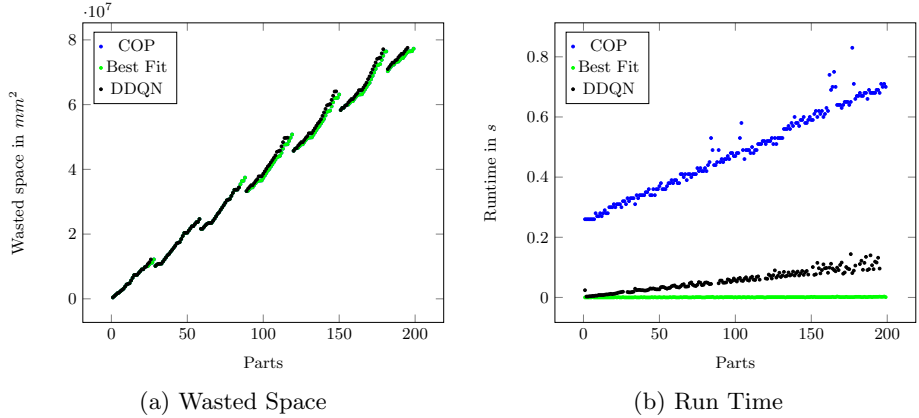
\begin{figure*}
\begin{subfigure}[t]{0.5\textwidth}
\begin{tikzpicture} [scale=0.7]
	\begin{axis}[legend pos=north west, no markers,
	xlabel={Parts},
	ylabel={Wasted space in $mm^2$}]
		\addplot[only marks, mark size=0.75pt,blue] table [col sep=comma] {data/MiniZinc_Gurobi_wasted_space_sum4.csv};
		\addlegendentry{COP}
		\addplot[only marks, mark size=0.75pt,green] table [col sep=comma] {data/Best_Fit_wasted_space_sum4.csv};
		\addlegendentry{Best Fit}
		\addplot[only marks, mark size=0.75pt,black] table [col sep=comma] {data/DQN_wasted_space_sum4.csv};
		\addlegendentry{DDQN}
	\end{axis};	
\end{tikzpicture}
\caption{Wasted Space}
\end{subfigure}
~
\begin{subfigure}[t]{0.5\textwidth}
\begin{tikzpicture} [scale=0.7]
	\begin{axis}[legend pos=north west, no markers, 
	xlabel={Parts},
	ylabel={Runtime in $s$}]
	\addplot [only marks, mark size=0.75pt,blue] table [col sep=comma]{data/MiniZinc_Gurobi_run_time4.csv}; 
	\addlegendentry{COP}
	\addplot[only marks, mark size=0.75pt,green] table [col sep=comma] {data/Best_Fit_run_time4.csv};
	\addlegendentry{Best Fit}
	\addplot [only marks, mark size=0.75pt,black] table [col sep=comma] {data/DQN_run_time4.csv};
	\addlegendentry{DDQN}
	\end{axis};	
\end{tikzpicture}
\caption{Run Time}
\label{fig:Graphs:real}
\end{subfigure}
\label{fig:RealPartGraphs}
\caption{Graphs a and b show the results of the three evaluated algorithms  Constraint Programming (blue), Best Fit Heuristics (green), and DDQN (black) for real furniture parts.}
\end{figure*}

Fig. 4 and Fig. 5 show different graphs with the experimental results of the algorithm evaluation for 200 parts to be packed. Each data point in the graphs represents a sorting process for a certain number of parts. The graphs a to f in Fig. 4 show the solution quality and runtime of the algorithms on randomly generated parts. To make the randomly generated data deterministic, random seeds are used for its generation. For graphs a and b the random seed 1 is used. For graphs c and d as well as e and f the random seed is incremented by one. In graphs a and b of Fig. 5 the solution quality and runtime on real furniture parts are shown. Graph a in Fig. 5 shows certain gap\textit{}s between the data points at various points. These can be explained by the fact that after a certain number of parts, additional capacities are generated for sorting. Also the real furniture parts do not vary as much in length and width as randomly generated parts. Thus the tested algorithms can use better sorting strategies for real furniture parts with each extension of capacities. As a result, the wasted space can be reduced by additional capacities despite the increasing number of parts.

\paragraph{Investigating Hypothesis 1 \& 2:} 
Reviewing the runtime graphs on the right in Fig. 4 and 5, we argue that Hypotheses 1 and 2 hold. The Constraint Optimization approach (blue) is the slowest of the three algorithms, with runtimes starting at about 0.25 seconds to pack one part and rising to 0.7 seconds for packing 200 parts. The runtime of the DDQN approach (black) rises slower, starting at 0.003 seconds for packing one part. With an increasing number of parts, the runtime of DDQN starts to alternate between about 0.09 and 0.16 seconds for packing 180 parts. The runtime for the Best Fit heuristic (green) is barely measurable for packing one part and rises slowly to about 0.001 seconds for packing 200 parts.

\paragraph{Investigating Hypothesis 3:} 
In terms of solution quality (graphs on the left), the Deep Reinforcement Learner is not able to provide feasible solutions for all of the presented instances. Instead the results are scattered: In Graph e of Fig. 5 it can pack 11 parts and in Graph c it can pack 182 parts. Often the learner may choose a module that is too small if the part's length or width is close to the length or width of the next smaller module. So, instead of getting stuck in dead ends, the learner provides an infeasible solution because he has not yet approximated the module's dimensions accurately enough. This assumption is further supported by the experiment on real furniture parts, where the DDQN can store nearly all parts in the given number of trolleys because the real furniture parts are hardly in the border areas of the modules with their length and width.

On the contrary, the data points for the Best Fit Heuristic are not visible as they completely overlap with the results of COP. This means both approaches can solve the presented instances and provide equal solutions in terms of solution quality. In other words, the Best Fit Heuristic also provides an optimal solution in the instances considered. Therefore, we can conclude that Hypothesis 3 does not hold: While the DDQN approach does not provide feasible solutions in every configuration, this is rather due to the inaccurate approximation than a series of suboptimal decisions. 
Further, the Best Fit Heuristic generates feasible and optimal solutions, which we will investigate further in \cref{discussion}.

\subsection{Discussion}
\label{discussion}
\paragraph{Discussing the performance of DDQN:} 
Reviewing the solution quality graphs on the left, we can see that in some cases the DDQN approaches the COP. We found packing is not always optimal, i.e. the wasted space is slightly higher than the baseline. The aforementioned edge cases are the main reason for infeasible solutions. We suspect this is due to the training based on the Best Fit heuristic. A learner trained on the wasted space might instead choose a larger module and show the expected behavior of getting stuck in a dead-end due to a series of suboptimal choices. We argue that further training might improve the learner's approximation and cover those edge cases.  However, considering the good results of the Best Fit heuristic, further experiments were omitted. We argue that practitioners should prefer a heuristic or constraint optimization approach in this use case, as they provide optimal results regardless of training.

\paragraph{Discussing the applicability of the best fit heuristic:} 
Encouraged by the good results, we attempt to sketch a proof that the Best Fit heuristic always produces optimal results given the assumptions in \cref{problem_assumptions}. Suppose we have parts $P = \{1, \ldots, p\}$ and modules $M = \{1, \ldots, m\}$ with $|m| \geq |p|$. Every part and module have respective area sizes $\mathit{size}(p)$ and $\mathit{size}(m)$, respectively. We seek a feasable allocation $f : P \to M$ such that $\mathit{size}(f(m)) \geq \mathit{size}(m)$ holds for every module $m$. A best-fit for a part $p$, denoted by $\mathit{bf}(p)$, is given by the smallest available module that is still larger than $p$. We propose to allocate each part to its best-fit (available) module. This procedure guarantees an optimal allocation with respect to overall objective: $\mathrm{minimize} \sum_{p \in P}  \mathit{size}(f(p)) - \mathit{size}(p) $. 

Proof sketch: 
Let $p_i$ denote the parts. For the first part $p_1$, the best-fit choice optimizes the overall objective by definition. This also holds for the subsequent choice $p_2$, as trading modules $f'(p_1) = f(p_2)$ and $f'(p_2) = f(p_1)$ at best results in the same objective term $(\mathit{size}(f(p_1)) + \mathit{size}(f(p_2))) - (\mathit{size}(p_1) + \mathit{size}(p_2))$ for both allocations $f$ and $f'$, assuming that $f$ and $f'$ are feasable. Conversely, picking a larger than necessary module for $p_2$ only increases the objective value unnecessarily. Hence, we can not improve previous allocations. This argument can be repeated for subsequent parts.

Only one case is problematic: We allocate a slot in a module that is needed by a subsequent part. But this results in the instance being unsatisfiable, no matter in which order the parts are presented. Therefore, we conclude that under the assumptions made, the Best Fit heuristic generates an optimal solution for every satisfiable problem instance.

%% file: sections/Evaluation/ExperimentalSetup.tex
\newcounter{counter}
\setcounter{counter}{1}

\loop
\begin{subfigure}[t]{0.5\textwidth}
\begin{tikzpicture} [scale=0.7]
	\begin{axis}[legend pos=north west, no markers,
	xlabel={Parts},
	ylabel={Wasted space in $mm^2$}]
		\addplot[only marks, mark size=0.75pt,blue] table [col sep=comma] {data/MiniZinc_Gurobi_wasted_space_sum\arabic{counter}.csv};
		\addlegendentry{COP}
		\addplot[only marks, mark size=0.75pt,green] table [col sep=comma] {data/Best_Fit_wasted_space_sum\arabic{counter}.csv};
		\addlegendentry{Best Fit}
		\addplot[only marks, mark size=0.75pt,black] table [col sep=comma] {data/DQN_wasted_space_sum\arabic{counter}.csv};
		\addlegendentry{DDQN}
	\end{axis};	
\end{tikzpicture}
\caption{Wasted Space (Rand. Seed \arabic{counter})}
\end{subfigure}
~
\begin{subfigure}[t]{0.5\textwidth}
\begin{tikzpicture} [scale=0.7]
	\begin{axis}[legend pos=north west, no markers, 
	xlabel={Parts},
	ylabel={Runtime in $s$}]
	\addplot [only marks, mark size=0.75pt,blue] table [col sep=comma]{data/MiniZinc_Gurobi_run_time\arabic{counter}.csv}; 
	\addlegendentry{COP}
	\addplot[only marks, mark size=0.75pt,green] table [col sep=comma] {data/Best_Fit_run_time\arabic{counter}.csv};
	\addlegendentry{Best Fit}
	\addplot [only marks, mark size=0.75pt,black] table [col sep=comma] {data/DQN_run_time\arabic{counter}.csv};
	\addlegendentry{DDQN}
	\end{axis};	
\end{tikzpicture}
\caption{Run Time (Rand. Seed \arabic{counter})}
\label{fig:Graphs \arabic{counter}}
\end{subfigure}

\stepcounter{counter}
\ifnum \arabic{counter}<4
\repeat

%% file: sections/Conclusion.tex
\section{Conclusion}
\label{conclusion}

In this paper, we consider the problem of packing workpieces into a trolley in the furniture industry. 
We formulate the problem of assigning workpieces to trolley modules as an optimization problem, where the wasted space is minimized.
Our problem definition considers the rotation of workpieces and ensures that the capacity of the trolley and its modules are not violated.
To guide practitioners, we compare three approaches to solve the problem in terms of solution quality and runtime: 
While a constraint optimization approach acts as our baseline in terms of solution quality, the best fit heuristic and a deep reinforcement learning approach promise faster but possibly suboptimal results.

In our experiments on randomly generated and real-world data, both the optimization approach and the best fit heuristic achieve optimal results. 
We sketch proof to demonstrate that the best fit heuristic produces an optimal solution to our problem given that a solution exists.
Further, the best fit heuristic is faster than the constraint optimization approach solving instances with 200 workpieces in about 0.001 seconds compared to about 0.7 seconds.
Regarding the runtime, the deep reinforcement learning approach is located between the heuristic and the constraint optimization approach. 
It was not able to solve all given instances optimally and could not solve other instances at all. We assume this issue could be solved with more training but omitted further investigations because of the optimal results generated by the best fit heuristic. We conclude that practitioners can safely use the constraint optimization approach for the given problem as it produces optimal results.
However, our results indicate that the best fit heuristic works equally well in terms of solution quality while being faster. 
Therefore the use of the heuristic can also be recommended.
It is easy to implement and does not require the additional dependency of a constraint solver.
Lastly, deep reinforcement learning seems like a suitable, modern approach, but our results indicate that it is a bad fit for this use case, especially when the dimensions of workpieces are very diverse. 